\pdfoutput=1
\documentclass[10pt,twocolumn,letterpaper]{article}
\usepackage[pagenumbers]{cvpr}
\usepackage{times,epsfig,graphicx,amsmath,amssymb,amsthm,booktabs,bm,xcolor,multirow}
\newtheorem{theorem}{Theorem}
\newtheorem{proposition}[theorem]{Proposition}

\theoremstyle{remark}

\theoremstyle{plain}
\DeclareMathOperator{\tr}{tr}

\DeclareMathOperator{\erf}{erf}
\DeclareMathOperator{\wrap}{wrap}

\newcommand{\boxT}{\mathrm{box}_T}
\newcommand{\tbd}[1]{\textcolor{red}{#1}}

\newcommand{\nm}[1]{\ifcsname #1\endcsname\csname #1\endcsname\else\tbd{??}\fi}\def\stripplus#1#2\relax{\if+#1#2\else#1#2\fi}\newcommand{\nmu}[1]{\ifcsname #1\endcsname\expandafter\expandafter\expandafter\stripplus\csname #1\endcsname\relax\else\tbd{??}\fi}\def\stripsign#1#2\relax{\if-#1#2\else\if+#1#2\else#1#2\fi\fi}\newcommand{\nma}[1]{\ifcsname #1\endcsname\expandafter\expandafter\expandafter\stripsign\csname #1\endcsname\relax\else\tbd{??}\fi} %

\definecolor{cvprblue}{rgb}{0.21,0.49,0.74}
\usepackage[pagebackref,breaklinks,colorlinks,allcolors=cvprblue]{hyperref}
\usepackage[T1,OT1]{fontenc}
\definecolor{mnA}{HTML}{A61E4D}\definecolor{mnInk}{HTML}{1B1F2A}
\ifdefined\pdfmapline\pdfmapline{=plxSans-Bold-tlf-t1--base plxSans-Bold "AutoEnc_gmchnqiogedxb6eqrw52occsyg ReEncodeFont" <[plx_gmchnq.enc <plxSans-Bold.pfb}\fi
\DeclareFontFamily{T1}{obplex}{}\DeclareFontShape{T1}{obplex}{b}{n}{<-> plxSans-Bold-tlf-t1}{}
\newcommand{\mnLogo}{\mbox{\textcolor{mnInk}{\rule{1.7pt}{0.5em}}\hspace{0.06em}\textcolor{mnA}{\rule[0.22em]{1.7pt}{0.5em}}\hspace{0.06em}\textcolor{mnInk}{\rule{1.7pt}{0.5em}}\hspace{0.06em}\textcolor{mnA}{\rule[0.22em]{1.7pt}{0.5em}}\hspace{0.06em}\textcolor{mnInk}{\rule{1.7pt}{0.5em}}}}
\hypersetup{pdftitle={OffBeat: Multicoset Sampling for 4D Capture Beyond the Camera Frame Rate},pdfauthor={Chi-Wei Lee, An-Kai Chen, Cheng-Han Chiang, Yi-Chun Chang, Po-Chih Kuo}}
\def\confName{CVPR}
\def\confYear{2027}

\title{\mnLogo\hspace{0.3em}{\usefont{T1}{obplex}{b}{n}\textcolor{mnA}{Off}\textcolor{mnInk}{Beat}}: Multicoset Sampling for 4D Capture\\ Beyond the Camera Frame Rate}
\author{Chi-Wei Lee \quad An-Kai Chen \quad Cheng-Han Chiang \quad Yi-Chun Chang \quad Po-Chih Kuo\\
National Tsing Hua University, Hsinchu, Taiwan\\[2pt]
{\small\href{https://arthur031221.github.io/offbeat/}{\textcolor{mnA}{\texttt{arthur031221.github.io/offbeat}}}}}

\usepackage{iftex}
\ifPDFTeX\else
  \usepackage[no-math]{fontspec}
\fi

\newif\ifLadder\Ladderfalse     %
\newif\ifKrate\Kratefalse       %
\newif\ifCrit\Critfalse         %
\newif\ifPanSeeds\PanSeedsfalse %
\newif\ifBack\Backfalse         %
\newif\ifReal\Realfalse         %
\newcommand{\figph}[3]{\IfFileExists{#1}{\includegraphics[width=#2]{#1}}{\fbox{\parbox[c][#3][c]{\dimexpr#2-2\fboxsep-2\fboxrule\relax}{\centering\small figure pending}}}}

\begin{document}
\makeatletter\newcommand\teasercaption[1]{\def\@captype{figure}\caption{#1}}\makeatother
\twocolumn[{%
\renewcommand\twocolumn[1][]{#1}%
\maketitle
\begin{center}
\vspace{-1.6em}
{\setlength{\unitlength}{\textwidth}\begin{picture}(1,0.362319)%
\put(0,0){\includegraphics[width=\textwidth]{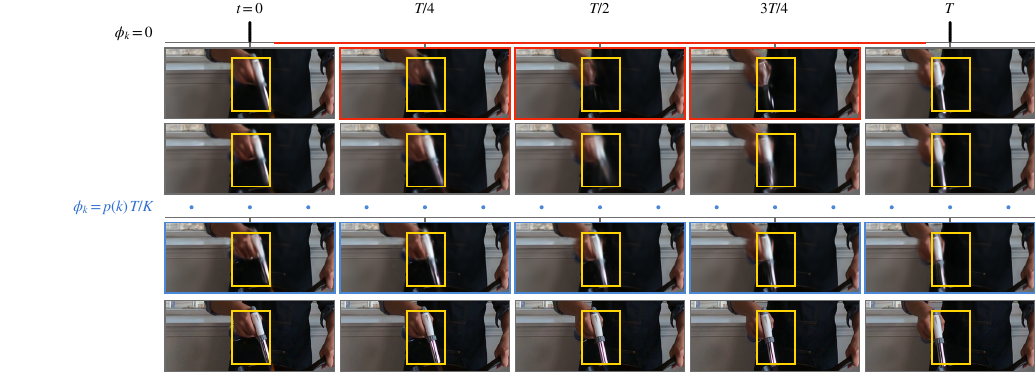}}%
\put(0.149275,0.281763){\makebox(0,0)[r]{\small\begin{tabular}{@{}r@{}}Synchronized\\STG~\cite{li2024spacetime}\end{tabular}}}%
\put(0.149275,0.209058){\makebox(0,0)[r]{\small\color[HTML]{444444}\begin{tabular}{@{}r@{}}STG~\cite{li2024spacetime} +\\interpolation~\cite{li2023amt}\end{tabular}}}%
\put(0.149275,0.113164){\makebox(0,0)[r]{\small\bfseries\color[HTML]{2F6FD0}\begin{tabular}{@{}r@{}}OffBeat\\(Ours)\end{tabular}}}%
\put(0.149275,0.037560){\makebox(0,0)[r]{\small Reference}}%
\end{picture}}
\teasercaption{\textbf{With the same cameras and frame count, staggered trigger times let OffBeat follow the motion between synchronized frames.} Each of the 18 training cameras of N3DV Flame Salmon keeps five frames of a 2\,s clip. Test-view renders over one camera frame interval $T$ and the trigger offsets $\phi_k$ of both rigs, where camera $k$ is in group $p(k)$ of $K{=}12$. Red: instants the synchronized rig misses, where synchronized STG fades the torch and interpolation blurs it. Yellow boxes: most dynamic region. STG: Spacetime Gaussians.}
\label{fig:main}
\end{center}
\vspace{0.4em}
}]

{\let\thefootnote\relax\footnotetext{\raggedright Emails: {\tt \{levi74108520963, kylechen.ankai, faidavid7, changg587\}@gmail.com, kuopc@cs.nthu.edu.tw}}}
\begin{abstract}
For 4D reconstruction, a synchronized multi-camera rig records all views at the same timestamps and misses the motion between them. Staggered triggers spread the same frames over more timestamps. However, when the cameras along each row fire one after another, as in 4DSloMo's schedule on the Neural 3D Video (N3DV) rig, the row acts as one moving camera that confounds depth with motion. We introduce OffBeat, which models a staggered rig as a spatially distributed multicoset sampler: shifted copies of one sampling grid. From the calibration alone, OffBeat derives camera groups that fire together and their firing order. Since one group records each timestamp, a short-lived primitive can fit an error its cameras share, such as a color offset, and flicker in novel views. We show that a primitive's opacity over time weights all groups equally, for every number of groups, only if it lasts at least one camera frame interval. On Spacetime Gaussians, a minimum lifetime approximates this bound, and a closed-form smoothed box of that length satisfies it exactly, with less flicker than a Gaussian in its place at nearly equal PSNR. In simulated staggered capture on N3DV, OffBeat outperforms the full 4DSloMo pipeline with video diffusion on every test frame, by 3.04\,dB in mean dynamic-region PSNR. Between the first and last synchronized frame, it outperforms the best synchronized method of each scene, with or without frame interpolation, by \nmu{NDSyncMint}\,dB in mean dynamic-region PSNR, while synchronized capture is better at its own frames. On PanopticSports, in this frame range, OffBeat is comparable in PSNR to synchronized capture with interpolation. On every N3DV scene, replacing 4DSloMo's schedule with ours raises the dynamic-region PSNR of Spacetime Gaussians, 4DSloMo without diffusion, and our reconstruction. Multi-camera rigs could follow faster motion at the same frame count by changing only when each camera fires.\looseness=-1 %
\end{abstract}

\section{Introduction}
\label{sec:intro}
\begin{figure*}[t]
\centering
\includegraphics[width=\linewidth]{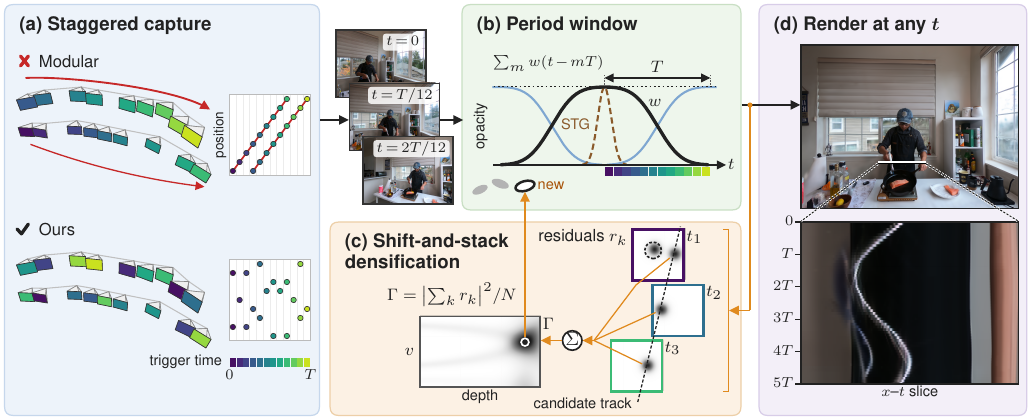}
\caption{\textbf{OffBeat overview.} Colors are trigger times within the camera frame interval~$T$. (a)~The modular schedule fires each camera row in position order, so the row acts as one moving camera, confounding depth with motion; ours keeps them separable. (b)~Each Gaussian's opacity follows a period window~$w$ whose copies $T$ apart (blue) sum to a constant, weighting all groups equally; STG's shorter Gaussian window (dashed) can fit one group's error. (c)~Residuals $r_k$ summed along a candidate track give a depth--velocity score $\Gamma$ that ranks a single-camera error (circled) low; new Gaussians are added at the maxima of $\Gamma$. (d)~Our test-view render and $x$--$t$ slice.\looseness=-1}
\label{fig:method}
\end{figure*}
4D reconstruction from multi-view video~\cite{li2022neural,li2024spacetime,wu20244d} still fails on fast non-rigid motion, such as a flame, a hand turning food, or a ball leaving a hand. One cause is temporal sampling: in a synchronized rig, all $N$ cameras record at the same timestamps, once every camera frame interval $T$. Adding cameras adds views of these timestamps and leaves the temporal bandwidth at $1/(2T)$. A higher frame rate costs resolution, exposure, storage and link bandwidth.\looseness=-1

Staggered triggers fire the cameras in $K$ groups spaced $T/K$ apart, which spreads the same number of frames over $K$ times as many timestamps (Fig.~\ref{fig:main}). For nearly co-located cameras this is the classic high-speed array~\cite{wilburn2005high,shechtman2005space,agrawal2010optimal}, and in a dense array staggered triggers sampled the video light field better than synchronized ones~\cite{wilburn2005high}. 4DSloMo~\cite{chen20254dslomo} staggers a multi-view rig to raise its effective frame rate and repairs sparse-view artifacts at each timestamp with a 14B-parameter video diffusion model.\looseness=-1

On the Neural 3D Video rig (N3DV)~\cite{li2022neural}, the modular schedule of 4DSloMo, which gives camera $k$ the offset $(k\bmod K)\,T/K$, fires each of the two camera rows in position order. Within a camera frame interval, such a row acts as one camera moving at constant speed: regardless of row size, a static point is indistinguishable from a moving point at another depth (Sec.~\ref{sec:sweep}, Fig.~\ref{fig:method}a).\looseness=-1 %

We model a staggered rig as a spatially distributed multicoset sampler~\cite{papoulis1977generalized,venkataramani2001optimal}: cameras that fire together form a \emph{trigger group} and sample at the camera frame rate on a shifted copy of one grid. As in multichannel sampling, a point whose motion is band-limited to half the effective frame rate $K/T$ is identifiable if every group triangulates it (Sec.~\ref{sec:theory}). Single-camera groups, as on N3DV, cannot triangulate; a motion model over one camera frame interval pools them. From the calibration alone, we choose the groups by how well the others triangulate, and their order by the predictive variance of novel views (Sec.~\ref{sec:schedule}).\looseness=-1

The reconstruction faces a second problem: as each timestamp is recorded by a single group, an error its cameras share, such as a color or calibration offset, is fit exactly by spurious content present only at those timestamps, which a novel view sees as flicker at the camera frame rate (Sec.~\ref{sec:confound}). A primitive can supply such content in every period only if its temporal opacity window, summed over copies one period apart, weights the groups unequally. We show that the windows weighting all groups equally for any number of groups and at any center are exactly the convolutions of a box of length $T$ (Prop.~\ref{prop:window}).
Hence a primitive's lifetime (its window's duration) is at least $T$ rather than the trigger spacing $T/K$, while its motion is sampled at the effective frame rate. On Spacetime Gaussians (STG)~\cite{li2024spacetime}, a \emph{lifetime floor} approximates this bound, and a closed-form \emph{period window}, a smoothed box, satisfies it exactly (Sec.~\ref{sec:window}).\looseness=-1 %

\begin{figure*}[t]
\centering
{\setlength{\unitlength}{\textwidth}\hypersetup{citecolor=black}\begin{picture}(1,0.295273)%
\put(0,0){\includegraphics[width=\textwidth]{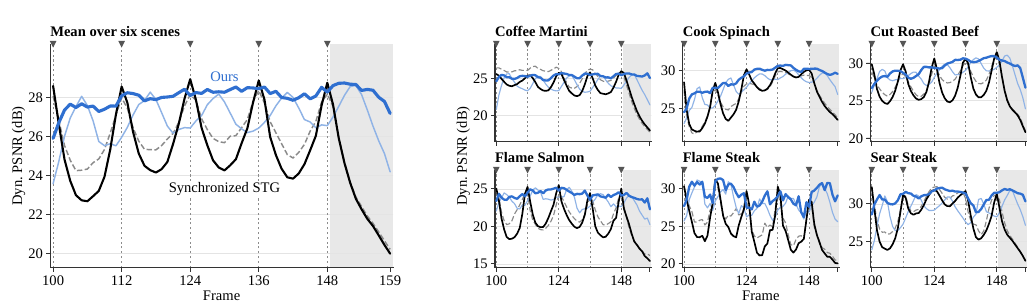}}%
\put(0.5,0.286545){\makebox(0,0){\footnotesize\begin{tabular}{@{}l@{\hspace{3pt}}l@{\hspace{10pt}}l@{\hspace{3pt}}l@{\hspace{10pt}}l@{\hspace{3pt}}l@{\hspace{10pt}}l@{\hspace{3pt}}l@{}}%
\textcolor[HTML]{2F6FD0}{\rule[1.4bp]{14bp}{1.5bp}} & \textcolor[HTML]{2F6FD0}{OffBeat (Ours)} & \textcolor[HTML]{8FB3E6}{\rule[1.8bp]{14bp}{0.7bp}} & OffBeat, modular schedule & \rule[1.65bp]{14bp}{1.0bp} & Synchronized STG~\cite{li2024spacetime} & \textcolor[HTML]{8A8A8A}{\rule[1.8bp]{2.6bp}{0.7bp}\hspace{1.2bp}\rule[1.8bp]{2.6bp}{0.7bp}\hspace{1.2bp}\rule[1.8bp]{2.6bp}{0.7bp}\hspace{1.2bp}\rule[1.8bp]{2.6bp}{0.7bp}} & STG~\cite{li2024spacetime} + interp.~\cite{li2023amt}%
\end{tabular}}}%
\end{picture}}
\caption{\textbf{Dynamic-region PSNR per frame on N3DV.} Left: mean over the six scenes. Right: each scene. Synchronized capture drops between its frames (\protect\raisebox{0.05em}{\protect\includegraphics[height=0.55em]{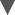}}) and after the last one (gray band), while OffBeat varies much less except near the first and last test frames.\looseness=-1}
\label{fig:perframe}
\end{figure*}
In summary, we make the following contributions:
\begin{list}{\textbullet}{\setlength{\leftmargin}{1em}\setlength{\labelwidth}{0.5em}\setlength{\labelsep}{0.5em}\setlength{\topsep}{0pt}\setlength{\partopsep}{0pt}\setlength{\itemsep}{0pt}\setlength{\parsep}{0pt}}
\item We introduce OffBeat, which models a staggered rig as a spatially distributed multicoset sampler, and show that a position-ordered row confounds depth with motion.
\item From the calibration alone, we derive which cameras fire together and in which order.
\item We show that temporal windows weighting all groups equally for every $K$ must last at least $T$, and derive a closed-form period window that satisfies this bound.
\item In simulated staggered capture on N3DV, OffBeat outperforms the full 4DSloMo pipeline and follows motion between synchronized frames better on average than a synchronized rig of equal camera frame rate, which is better at its own frames. On PanopticSports, up to the last synchronized frame, OffBeat is comparable in PSNR to synchronized capture with interpolation.\looseness=-1
\end{list} %

\section{Related Work}
\label{sec:related}
\paragraph{Staggered and unsynchronized capture.}
In high-speed arrays of nearly co-located cameras~\cite{wilburn2005high,shechtman2005space,agrawal2010optimal}, the firing order is a design parameter~\cite{wilburn2005high}. Color-coded illumination yields high-speed 3D under controlled light~\cite{novikov2026color}. Methods for unsynchronized video treat the time offsets as nuisances under smoothness priors~\cite{avidan2000trajectory,park20103d,vo2016spatiotemporal,zheng2015sparse}, estimate them~\cite{caspi2002spatio,albl2017two,kim2024sync,xu2026unsync,synctrack4d2025,zhu2026astra}, or recover structure and motion from asynchronous tracks~\cite{huang2025trajectory,su2025linear}. 4DSloMo~\cite{chen20254dslomo} staggers a rig with a modular schedule.\looseness=-1
\paragraph{Multichannel sampling.}
In multichannel sampling, $K$ channels at $1/K$ of the Nyquist rate recover a band-limited signal~\cite{papoulis1977generalized,venkataramani2001optimal}. Parallel MRI shares this structure~\cite{pruessmann1999sense,tsao2003kt}, and CAIPIRINHA~\cite{breuer2005caipirinha} shifts the sampling pattern such that aliases fall where the coil sensitivities differ. Our ordering criterion plays this role for cameras by decorrelating the trigger offsets from the camera positions.\looseness=-1
\paragraph{4D Gaussians and sampling-aware representations.}
Dynamic extensions of Gaussian splatting~\cite{kerbl20233d} use deformation fields~\cite{wu20244d}, native 4D primitives~\cite{yang2024real}, per-Gaussian trajectories with temporal opacity~\cite{li2024spacetime,lee2024fully,wang2025freetimegs}, or tracked Gaussians~\cite{luiten2024dynamic}. RetimeGS~\cite{wang2026retimegs} and SPIN-4DGS~\cite{kim2026spin} target temporal aliasing and large displacements between synchronized frames. Mip-Splatting~\cite{yu2024mip} bounds the 3D frequency of a primitive by the sampling rate of the training views, and our lifetime bound is its temporal counterpart. SharpTimeGS~\cite{liao2026sharptimegs} learns a flat-top lifespan for each primitive to balance static and dynamic content, whereas our window's box has the derived length $T$.\looseness=-1

\section{Method}
\subsection{Staggered rigs as multicoset samplers}
\label{sec:theory}
\label{sec:setup}
Each of the $N$ training cameras records a frame once every camera frame interval $T$. Camera $k$ fires at its trigger offset $\phi_k$, and a synchronized rig has all offsets zero. A $K$-group stagger fires the cameras $S_p$ of group $p$ at the offset $pT/K$. All triggers form the \emph{joint grid} of spacing $T/K$, whose rate $K/T$ is the effective frame rate of the rig~\cite{chen20254dslomo}. Each instant of this grid is recorded by one group alone.\looseness=-1

Camera $k$ observes a scene point $x$ through its $2\times3$ projection Jacobian $D_k$, with Fisher information $G_k=D_k^{\top}D_k/\sigma^2$ for pixel noise $\sigma$. $G_k$ has rank two, and its null space is the viewing ray $\ell_k$. Group $p$ has the information $H_p=\sum_{k\in S_p}G_k$. A single group cannot separate frequencies that differ by a multiple of the camera frame rate. The $K$ such frequencies inside the band of the joint grid form an \emph{alias set}, and the Fourier coefficients of $x(t)$ there are its \emph{alias components}; each group observes its own phase-weighted sum of them. For a trajectory band-limited to the Nyquist frequency of the joint grid, the Fisher information of its Fourier coefficients is block-diagonal over alias sets, with blocks built from the $H_p$, as in multichannel sampling~\cite{papoulis1977generalized,pruessmann1999sense}.\looseness=-1

We prove two properties of this structure in Sec.~S1 of the supplementary material. First, in a synchronized rig the data depend on the trajectory only through alias sums, for any number of cameras, so adding synchronized cameras leaves the temporal bandwidth at $1/(2T)$. Second, in a $K$-group stagger, a trajectory with $K$ alias components per set is identifiable if and only if every group triangulates, $H_p\succ0$, \ie has two cameras with non-parallel rays to the point. When every group triangulates, staggering thus raises the temporal bandwidth $K$-fold. With $N<2K$, some groups have one camera and cannot triangulate.\looseness=-1

A motion model over one camera frame interval pools all groups. Let a point's position over this period $T$ be a degree-$J$ polynomial in time $d$ with coefficients $\theta\in\mathbb{R}^{3(J+1)}$, and let camera $k$ see it at $d_k$. The Fisher information of $\theta$ is then $\sum_k\mathbf{b}(d_k)\mathbf{b}(d_k)^{\top}\otimes G_k$ with $\mathbf{b}(d)=(1,d,\dots,d^J)$. For $J\geq1$, it is singular in a synchronized rig and nonsingular once $J{+}1$ groups triangulate (Sec.~S1).
A polynomial per period replaces the $K$ alias components by $J{+}1$ shared coefficients, as in trajectory triangulation~\cite{avidan2000trajectory,park20103d} and spatiotemporal bundle adjustment~\cite{vo2016spatiotemporal}.\looseness=-1

\begin{figure}[t]
\centering
{\setlength{\unitlength}{\linewidth}\hypersetup{citecolor=black}\begin{picture}(1,0.406790)%
\put(0,0){\includegraphics[width=\linewidth]{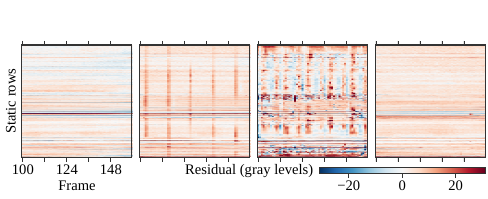}}%
\put(0.157827,0.326667){\makebox(0,0)[b]{\footnotesize\begin{tabular}[b]{@{}c@{}}Synchronized\\STG~\cite{li2024spacetime}\end{tabular}}}%
\put(0.399276,0.326667){\makebox(0,0)[b]{\footnotesize\begin{tabular}[b]{@{}c@{}}Staggered\\STG~\cite{li2024spacetime}\end{tabular}}}%
\put(0.640724,0.326667){\makebox(0,0)[b]{\footnotesize\begin{tabular}[b]{@{}c@{}}4DSloMo~\cite{chen20254dslomo}\\recon.\ stage\end{tabular}}}%
\put(0.882173,0.326667){\makebox(0,0)[b]{\footnotesize\color[HTML]{2F6FD0}\begin{tabular}[b]{@{}c@{}}OffBeat\\(Ours)\end{tabular}}}%
\end{picture}}%
\par\vspace{3pt}\includegraphics[width=\linewidth]{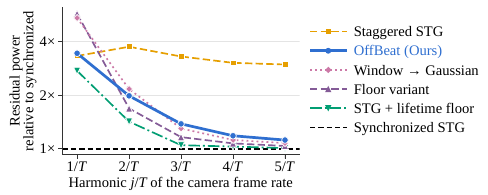}
\caption{\textbf{Flicker on static pixels.} Top: test-view residual over test frames on Flame Salmon along the image column at 1/4 of the width. Ticks mark synchronized frames; vertical stripes show static pixels that change together. Bottom: static-pixel residual power at the harmonics $j/T$ relative to synchronized STG, geometric mean over the six N3DV scenes. Staggered methods use the OffBeat schedule. Floor variant: STG with a lifetime floor, per-image timestamps and no opacity reset. Window~$\to$~Gaussian: OffBeat with STG's Gaussian window and unscaled velocity.\looseness=-1}
\label{fig:flicker}
\end{figure}
\subsection{Trigger schedule}
\label{sec:schedule}
\begin{table*}[t]
\centering
\caption{\textbf{Comparison on N3DV} ($K{=}12$, 2.5\,fps per camera, test frames 100--159). Dyn.\ PSNR (dynamic regions, Sec.~\ref{sec:exp_setup}) is also given over frames 100--148, where synchronized capture does not extrapolate. Best in bold, second best underlined.\looseness=-1}
\label{tab:main}
\setlength{\tabcolsep}{5pt}\renewcommand{\arraystretch}{0.95}
\footnotesize
\begin{tabular}{l cccccc cccc}
\toprule
& \multicolumn{6}{c}{Dynamic-region PSNR$\uparrow$ (dB)} & \multicolumn{4}{c}{Mean over scenes}\\
\cmidrule(lr){2-7}\cmidrule(lr){8-11}
Method & Coffee & Cook & Beef & F.\,Salmon & F.\,Steak & Sear & \multicolumn{2}{c}{Dyn.\ PSNR$\uparrow$} & PSNR$\uparrow$ & LPIPS$\downarrow$\\
\cmidrule(lr){8-9}
& & & & & & & 100--159 & 100--148 & & \\
\midrule
\multicolumn{11}{l}{\textit{Synchronized capture}} \\
STG~\cite{li2024spacetime} & 23.17 & 25.88 & 25.95 & 19.98 & 23.84 & 27.16 & 24.33 & 24.97 & 30.57 & .117 \\
STG~\cite{li2024spacetime} + interpolation~\cite{li2023amt} & 23.88 & 25.95 & 26.57 & 20.59 & 24.79 & 27.91 & 24.95 & 25.85 & 30.78 & .129 \\
\midrule
\multicolumn{11}{l}{\textit{Staggered capture, modular schedule}} \\
4DSloMo~\cite{chen20254dslomo}, recon.\ stage & 18.86 & 24.44 & 24.49 & 19.34 & 23.97 & 25.32 & 22.74 & 22.74 & 25.26 & .218 \\
4DSloMo~\cite{chen20254dslomo}, full pipeline & 22.79 & 25.26 & 26.75 & 21.61 & 25.46 & 27.27 & 24.86 & 24.95 & 26.90 & .186 \\
\midrule
\multicolumn{11}{l}{\textit{Staggered capture, OffBeat schedule}} \\
4DSloMo~\cite{chen20254dslomo}, recon.\ stage & 21.20 & 26.49 & 26.36 & 21.63 & 27.08 & 28.20 & 25.16 & 25.14 & 27.11 & .176 \\
STG~\cite{li2024spacetime} & \underline{24.94} & 27.87 & 28.24 & 23.69 & 27.73 & 30.08 & 27.09 & 27.05 & \underline{30.85} & .116 \\
STG~\cite{li2024spacetime}, 2$\times$ iterations & 24.65 & \underline{28.14} & \underline{28.80} & \underline{23.77} & \underline{28.06} & \underline{30.79} & \underline{27.37} & \underline{27.31} & 30.79 & \underline{.108} \\
\textbf{OffBeat (Ours)} & \textbf{25.26} & \textbf{28.76} & \textbf{29.25} & \textbf{24.16} & \textbf{29.04} & \textbf{30.92} & \textbf{27.90} & \textbf{27.83} & \textbf{31.25} & \textbf{.104} \\
\bottomrule
\end{tabular}
\end{table*}
\paragraph{Position-ordered triggers.}\label{sec:sweep}
The trigger order leaves the identifiability condition of Sec.~\ref{sec:theory} unchanged but can undo the gain of staggering. Let cameras with centers $o_k=o_0+s_ke$ on a line each fire once, at $t_k=t_0+\alpha s_k$ with $\alpha\neq0$, and let a point move as $X(t)=X_0+v(t-t_0)$. For every $\lambda>0$, the point with $X_0'=o_0+\lambda(X_0-o_0)$ and $v'=\lambda v+(1-\lambda)e/\alpha$ has the same image in every camera. The Fisher information of $(X_0,v)$ from one such sweep is singular for any number of cameras (proof in Sec.~S3).
The sweep thus confounds depth with motion, turning parallax into apparent motion (Fig.~\ref{fig:method}a). This is the ambiguity of a camera and a point both moving at constant velocity~\cite{nardone1981observability,avidan2000trajectory,park20103d}, and of a slowly moving camera~\cite{gao2022monocular}; in a rig, the schedule decides whether it arises. Repeated sweeps resolve it only for points keeping their velocity over several periods, unlike fast non-rigid motion. On N3DV, where the modular schedule sweeps both rows (Sec.~S3), only their vertical offset keeps the information nonsingular.\looseness=-1 %

\paragraph{Grouping.}\label{sec:twolevel} The partition into groups sets the information $H_p$ of each instant, on which identifiability depends. With N3DV's 17 to 20 training cameras in 12 groups, at least four groups have a single camera under any partition, and the motion model pools them. Over the other groups, we maximize $\mathrm{mean}_p\lambda_p+\frac12\min_p\lambda_p$, where $\lambda_p$ is the working-volume average of the smallest eigenvalue of $H_p$ (Sec.~S3).\looseness=-1

\paragraph{Ordering.} For a point moving at constant velocity $v$ over one period centered at $\mu$ and seen once per camera, pooling with $J{=}1$ gives the information of $(x,\,vT/2)$,\looseness=-1
\begin{equation}
\mathcal{I}(\mu)=\sum_k\begin{bmatrix}1&d_k\\ d_k&d_k^2\end{bmatrix}\otimes G_k,\qquad d_k=\frac{\wrap(\phi_k-\mu)}{T/2},
\label{eq:fisher}
\end{equation}
where $d_k\in[-1,1)$ is the wrapped trigger offset from $\mu$ in half periods. For novel views, we minimize $\max_{u,t}V+\frac12\,\mathrm{mean}_{u,t}V$ with the predictive variance\looseness=-1
\begin{equation}
V(u,t)=\mathbb{E}_{\mu,x}\tr\big(Q_u\,A\,\mathcal{I}(\mu)^{-1}A^{\top}\big),\quad A=[I,\;\tau I],
\label{eq:pv}
\end{equation}
where $\tau$ is the wrapped offset of a trigger instant $t$ from $\mu$ in half periods, $Q_u$ is the information of a virtual view $u$ midway between neighboring training cameras, and the expectation is over centers $\mu$ in one period and working-volume points $x$. Grouping and ordering follow the E- and G-optimality criteria of optimal design~\cite{kiefer1974general,pukelsheim2006optimal}, respectively. A local search optimizes them in seconds on a CPU (Alg.~S1). Both need only the calibration; their weights $\frac12$ hardly affect $\max V$ (Sec.~S3). The modular schedule's $\max V$ per N3DV scene is 2.2--2.4 times ours.\looseness=-1 %

\subsection{Lifetime bound}
\label{sec:method}
\paragraph{Camera--time confound.}\label{sec:confound}
Any error common to a group's views, such as a calibration or exposure error of a single-camera group, can be explained exactly by content present only when that group fires. If each group has its own error at a pixel, such as a color offset, the periodic signal $h$ that equals the firing group's error reproduces it in every training image. In a synchronized rig, $h$ is constant. A test camera belongs to no group and sees $h$ as flicker at the camera frame rate and its harmonics (Fig.~\ref{fig:flicker}, Sec.~\ref{sec:exp_rule2}). On N3DV, one or two cameras see each instant, so this explanation is always available.\looseness=-1

\paragraph{Window condition.}\label{sec:wincond}
Primitive $i$ with temporal center $\mu_i$ and window $w(t-\mu_i)$, such as STG's temporal opacity~\cite{li2024spacetime}, gives group $p$ the total weight\looseness=-1
\begin{equation}
\Omega_p(\mu_i)=\sum_{m}w\Big(\tfrac{pT}{K}+mT-\mu_i\Big)=(\Pi_Tw)\Big(\tfrac{pT}{K}-\mu_i\Big),
\label{eq:period}
\end{equation}
where $\Pi_Tw$ sums the copies of $w$ shifted by whole periods (Fig.~\ref{fig:method}b). If $\Pi_Tw$ varies, primitives can, for some $K$, weight the groups unequally and supply $h$. Prop.~\ref{prop:window} (proof in Sec.~S1) characterizes the windows that prevent this.\looseness=-1
\begin{proposition}\label{prop:window}
For a compactly supported window $w$, the following are equivalent:
(a) $\Omega_p(\mu)$ is independent of $p$ for every center $\mu$ and every $K$.
(b) $\hat w(j/T)=0$ for every integer $j\neq0$, with $\hat w$ the Fourier transform of $w$.
(c) For any centers $\mu_i$ and amplitudes $b_i$, $\Pi_T\sum_ib_iw(\cdot-\mu_i)$ is constant, so no set of primitives weights the groups unequally.
(d) $w=\boxT*g$ for a compactly supported distribution $g$, with $\boxT=\mathbf{1}_{[-T/2,T/2)}$.
No such window is shorter than $T$, and the shortest ones are shifted multiples of $\boxT$.\looseness=-1
\end{proposition}
Prop.~\ref{prop:window} thus bounds a primitive's lifetime below by $T$ but leaves its motion sampled at $K/T$. Condition~(b) is the partition-of-unity or Strang--Fix condition~\cite{strang1973fourier} at scale $T$, and (d) is the order-one case of the B-spline factorization~\cite{blu2001moms}. The condition constrains the weights summed over all periods, over which a persistent single-camera error recurs, and leaves those within one period free.\looseness=-1

\subsection{Reconstruction on STG}
\label{sec:window}
\begin{figure*}[t]
\centering
{\setlength{\unitlength}{\textwidth}\begin{picture}(1,0.531250)%
\put(0,0){\includegraphics[width=\textwidth]{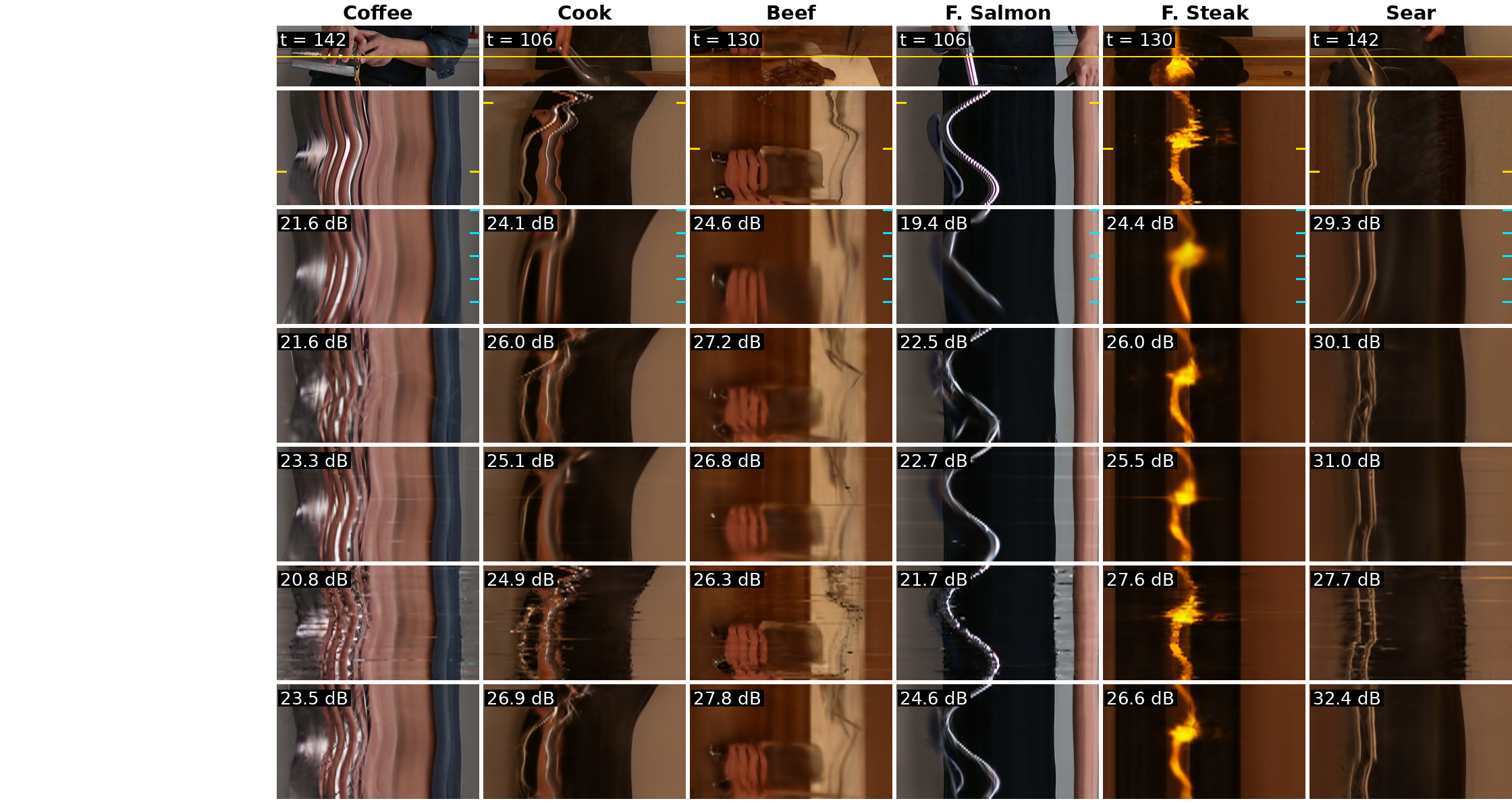}}%
\put(0.178571,0.494196){\makebox(0,0)[r]{\small Reference strip}}%
\put(0.178571,0.433482){\makebox(0,0)[r]{\small Reference slice}}%
\put(0.178571,0.354911){\makebox(0,0)[r]{\small\begin{tabular}{@{}r@{}}STG~\cite{li2024spacetime} + interp.~\cite{li2023amt}\\(synchronized)\end{tabular}}}%
\put(0.178571,0.276339){\makebox(0,0)[r]{\small\begin{tabular}{@{}r@{}}OffBeat\\(modular schedule)\end{tabular}}}%
\put(0.178571,0.197768){\makebox(0,0)[r]{\small\begin{tabular}{@{}r@{}}STG~\cite{li2024spacetime}\\(OffBeat schedule)\end{tabular}}}%
\put(0.178571,0.119196){\makebox(0,0)[r]{\small\begin{tabular}{@{}r@{}}4DSloMo~\cite{chen20254dslomo} recon.\\(OffBeat schedule)\end{tabular}}}%
\put(0.178571,0.040625){\makebox(0,0)[r]{\small\bfseries OffBeat (Ours)}}%
\end{picture}}
\caption{\textbf{\boldmath$x$--$t$ slices on N3DV} (Neural 3D Video). Top: strip at the yellow tick, midway between synchronized frames (cyan). Bottom: slices along its yellow line, time downward; slice PSNR in dB. Per scene, we pick the strip, instant and line of largest ground-truth change.\looseness=-1}
\label{fig:qual}
\end{figure*}
\begin{figure*}[t]
\centering
{\setlength{\unitlength}{\textwidth}\begin{picture}(1,0.274203)%
\put(0,0){\includegraphics[width=\textwidth]{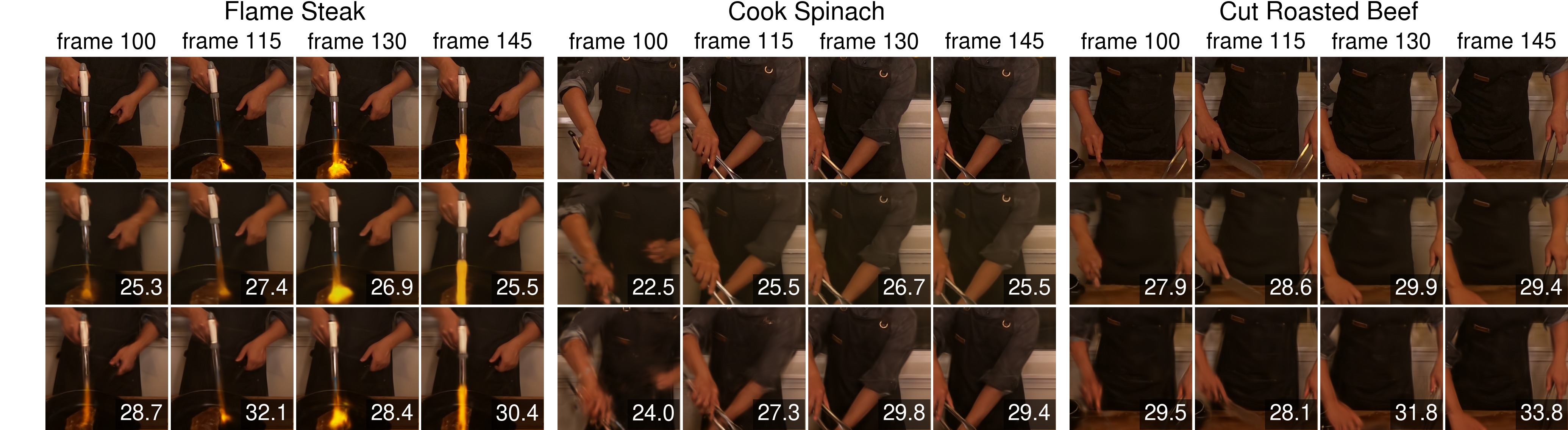}}%
\put(0.014493,0.198889){\makebox(0,0){\rotatebox{90}{\scriptsize Reference}}}%
\put(0.014493,0.118986){\makebox(0,0){\rotatebox{90}{\scriptsize 4DSloMo~\cite{chen20254dslomo}}}}%
\put(0.014493,0.039083){\makebox(0,0){\rotatebox{90}{\scriptsize\bfseries Ours}}}%
\end{picture}}
\caption{\textbf{OffBeat against the full 4DSloMo pipeline and schedule.} Crops around the $x$--$t$ slice lines, every 15 frames; crop PSNR in dB. OffBeat leads 4DSloMo in Dyn.\ PSNR and PSNR on every test frame of all N3DV scenes, but not always in crop PSNR (Sec.~S5).\looseness=-1}
\label{fig:slomofull}
\end{figure*}
We build on STG~\cite{li2024spacetime} and keep its loss and renderer. Each STG primitive $i$ has a temporal center $\mu_i$ and scale $a_i$ with temporal opacity $\exp(-(d/a_i)^2)$ at $d=t-\mu_i$, polynomial motion and rotation, and time-dependent features.\looseness=-1

\paragraph{Lifetime floor.}
Gaussian windows violate Prop.~\ref{prop:window}(b). Wide ones have little power at the harmonics $j/T$ (Sec.~S1), but narrow ones can favor one group. Rendering with the temporal scale $\sqrt{a_i^2+a_{\min}^2}$ in place of $a_i$, with $a_{\min}=T/4$, passes at most 29\% of the power at the camera frame rate and under 1\% at higher harmonics.\looseness=-1

\paragraph{Period window.}
Since $\Pi_T\boxT=1$, Prop.~\ref{prop:window}(d)$\Rightarrow$(c) holds for any integrable $g$, and a Gaussian $g$ gives
\begin{equation}
w_i(t)=\frac{\erf\big((d+\frac T2)/a_i\big)-\erf\big((d-\frac T2)/a_i\big)}{2\,\erf\big(T/(2a_i)\big)}.
\label{eq:window}
\end{equation}
For every center and scale, the period window meets the condition with no added parameter, gradients flow through the closed form, and it tends to the box window as $a_i\to0$. We keep the floor on $a_i$ because the window's gradient with respect to its center concentrates at the edges, and the floor spreads it from one to most of the 12 groups (Sec.~S1).\looseness=-1

Polynomial motion inside a window can reintroduce group-dependent components of $h$ that Prop.~\ref{prop:window}(c) excludes for constant amplitudes, so we rescale STG's velocity to the displacement over half a window (Sec.~S2).\looseness=-1 %

\paragraph{Training.}
\label{sec:train}
Two settings keep any update from depending on one group alone. With \emph{per-image timestamps}, the timestamp is drawn per image, not per batch as in STG. With \emph{no opacity reset}, the few cameras of one timestamp never have to re-estimate all opacities alone. We call STG with the floor and these two settings the \emph{floor variant}.\looseness=-1

\paragraph{Densification.}
\label{sec:birth}
Every training camera fires once in the period centered at $t_0$. For a small opaque primitive moving along $x+v(t-t_0)$ with a period window centered at $t_0$, let $r_k$ be the residual at the pixel it covers in camera $k$ when this camera fires. For a locally constant render, the primitive's best color lowers the squared error by $\Gamma(x,v,t_0)=\big|\sum_kr_k\big|^2/N$.
This is the shift-and-stack statistic of track-before-detect~\cite{reed1983tbd} and, to first order in the opacity, the dual certificate of conditional gradient on measures~\cite{bredies2013inverse,boyd2017adcg}. A residual $r$ shared by $m$ cameras scores $m^2|r|^2/N$, so errors common to one group (Sec.~\ref{sec:confound}) rank low (Fig.~\ref{fig:method}c). Instead of STG's guided sampling along high-error rays, we add primitives at the maxima of $\Gamma$ over depth and lateral velocity (Alg.~S2), keeping STG's cloning and splitting. Other methods densify by gradients or errors~\cite{kerbl20233d,rotabulo2024revising,wang2025steepgs,lyu2024resgs}, temporal visibility~\cite{sandu2026temporally}, or appearance change~\cite{roh2026dusting}.\looseness=-1

\section{Experiments}
\label{sec:exp}
\subsection{Setup}
\label{sec:exp_setup}
\paragraph{Data and protocol.}
N3DV~\cite{li2022neural} contains six dynamic scenes. We hold out camera~0 for testing and train on the other 17--20 cameras. Following 4DSloMo~\cite{chen20254dslomo}, we subsample each 30\,fps video to every 12th frame (2.5\,fps), with a separate offset per camera for staggered capture and a common one for synchronized capture. We report test frames 100--159 unless stated otherwise and validate on frames 40--99 of Flame Steak and Cook Spinach. PanopticSports~\cite{joo2015panoptic,luiten2024dynamic} has 27 training and 4 test cameras, and we use six sequences with test frames 0--59 and $K{=}12$, validating on frames 90--149 of Softball and Basketball. \looseness=-1 Each method is trained once per scene, and Sec.~S6 reports three seeds. %
\paragraph{Compared methods.}
For synchronized capture, we train STG and \emph{STG + interpolation}, STG trained on videos in which AMT~\cite{li2023amt} fills the 11 frames between synchronized frames. RetimeGS~\cite{wang2026retimegs}, for which we found no public code, reports this recipe with FILM~\cite{reda2022film} as its strongest N3DV baseline, 0.05\,dB below its own result. %
For staggered capture, we compare OffBeat with STG, also trained with twice the iterations, and with 4DSloMo~\cite{chen20254dslomo}, using its official code and settings. We run its reconstruction stage, a native 4D Gaussian model~\cite{yang2024real}, on both schedules and on the synchronized frames of N3DV (where it is weaker than STG); we run its full pipeline with video diffusion on all six N3DV scenes (Sec.~S5). All STG-based methods share the initialization, renderer and clock alignment (Sec.~S8) and are trained for 10k iterations unless stated otherwise. OffBeat trains in 423--481\,s per scene on one RTX~4090.\looseness=-1 %
\paragraph{Metrics.}
At the test view, we report PSNR, AlexNet LPIPS~\cite{zhang2018unreasonable} and dynamic-region PSNR (Dyn.\ PSNR, as in FreeTimeGS~\cite{wang2025freetimegs}); Sec.~S6 reports SSIM. Dynamic pixels are those whose ground-truth luma has a temporal standard deviation above 0.05 over the 60 test frames, dilated by 5\,px. Per scene and metric, the best synchronized method is the strongest of STG, STG + interpolation, the floor variant and, on N3DV, our reconstruction, all from synchronized capture. Sec.~S6 gives per-scene results with paired bootstrap intervals~\cite{kunsch1989jackknife}. Over 20 pairs of mask threshold and dilation, OffBeat's Dyn.\ PSNR difference keeps its sign against synchronized STG, STG + interpolation, STG on our schedule (except on PanopticSports) and the floor variant (negative on PanopticSports; Sec.~S6).\looseness=-1 %

\subsection{N3DV}
\label{sec:exp_main}
\begin{figure*}[t]
\centering
{\setlength{\unitlength}{\textwidth}\begin{picture}(1,0.288210)%
\put(0,0){\includegraphics[width=\textwidth]{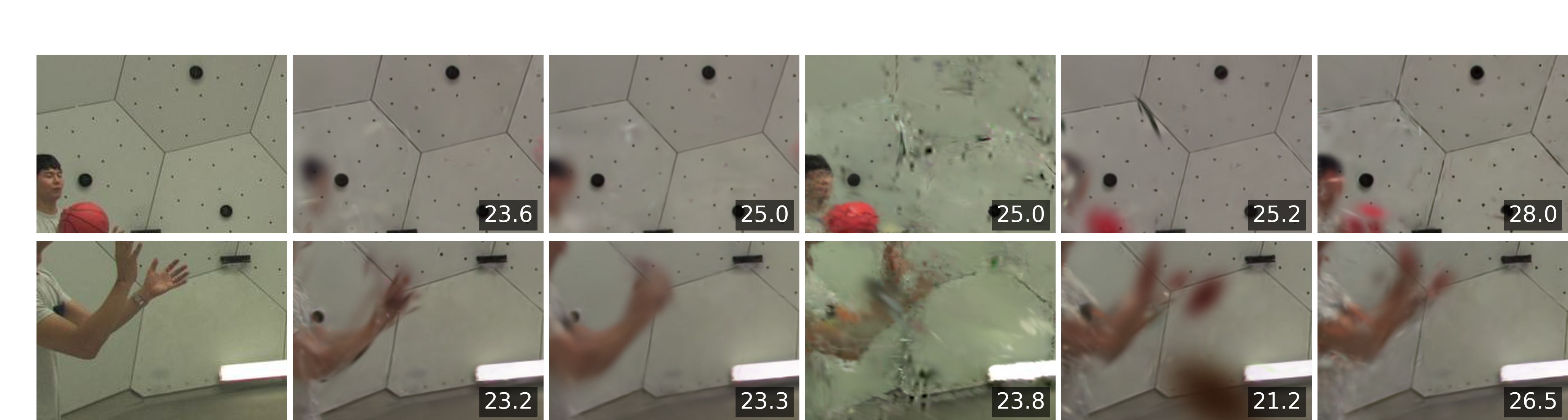}}%
\put(0.348242,0.279210){\makebox(0,0){\scriptsize\itshape Synchronized capture}}%
\put(0.756727,0.279210){\makebox(0,0){\scriptsize\itshape Staggered capture}}%
\put(0.189700,0.269710){\line(1,0){0.317100}}%
\put(0.516500,0.269710){\line(1,0){0.480500}}%
\put(0.103152,0.250771){\makebox(0,0){\scriptsize Reference}}%
\put(0.266545,0.250771){\makebox(0,0){\scriptsize STG~\cite{li2024spacetime}}}%
\put(0.429939,0.250771){\makebox(0,0){\scriptsize\begin{tabular}{@{}c@{}}STG~\cite{li2024spacetime} +\\interpolation~\cite{li2023amt}\end{tabular}}}%
\put(0.593333,0.250771){\makebox(0,0){\scriptsize\begin{tabular}{@{}c@{}}4DSloMo~\cite{chen20254dslomo}\\recon.\ stage$^\dagger$\end{tabular}}}%
\put(0.756727,0.250771){\makebox(0,0){\scriptsize STG~\cite{li2024spacetime}}}%
\put(0.920121,0.250771){\makebox(0,0){\scriptsize\bfseries OffBeat (Ours)}}%
\put(0.011636,0.176260){\makebox(0,0){\rotatebox{90}{\footnotesize Basketball}}}%
\put(0.011636,0.057056){\makebox(0,0){\rotatebox{90}{\footnotesize Football}}}%
\end{picture}}
\caption{\textbf{PanopticSports, between synchronized frames.} Synchronized capture loses the ball (Basketball) and smears the hands (Football). 4DSloMo's reconstruction stage renders the ball sharply but clutters the wall, and STG on our schedule adds dark streaks and blobs. OffBeat has the highest crop PSNR (dB), but the ball and hands stay blurred. Crops, frames and the sequences shown are where OffBeat leads all other columns by the most, among the 10\% of crops with the largest ground-truth change from the previous frame. $^\dagger$The better schedule in Dyn.\ PSNR. Sec.~S6 shows Softball and Tennis, where 4DSloMo's reconstruction stage is better in Dyn.\ PSNR.\looseness=-1}
\label{fig:pan_qual}
\end{figure*}
\paragraph{Staggered against synchronized capture.}
Between the first and last synchronized frame, OffBeat outperforms the best synchronized method by \nmu{NDSyncMint}\,dB in mean Dyn.\ PSNR (Table~\ref{tab:main}, Fig.~\ref{fig:perframe}) and on five of six scenes (Sec.~S6). At the five synchronized frames, the better of STG and STG + interpolation in each scene is 1.23\,dB above OffBeat in Dyn.\ PSNR but 2.20\,dB below on the 44 frames in between (Sec.~S6). %
In Fig.~\ref{fig:qual}, interpolation flattens the traces, the modular schedule fragments them, and STG on our schedule adds single-frame streaks. 4DSloMo's reconstruction stage is sharp but jitters, and its slice PSNR on Flame Steak is higher than ours. OffBeat follows the trajectories.\looseness=-1 %

\paragraph{Training length and camera frame rate.}
Synchronized STG trained twice as long remains \nmu{NXOursDStwoM}\,dB below OffBeat in Dyn.\ PSNR (Sec.~S6). With the official 30k iterations for both, OffBeat outperforms STG + interpolation between the first and last synchronized frame by 2.29\,dB in Dyn.\ PSNR on N3DV and by 0.99\,dB on PanopticSports (Sec.~S8). With twice the camera frame rate (5\,fps), synchronized STG is still 1.02\,dB below OffBeat at 2.5\,fps between its first and last synchronized frame. At 10\,fps it is 0.36\,dB better there, and at 30\,fps it is 0.60\,dB better over all test frames, or 1.20\,dB if it alone gets 30k iterations (Sec.~S6).\looseness=-1 %

\paragraph{Schedule.}
With our reconstruction, replacing our schedule by the modular one lowers Dyn.\ PSNR by \nmu{CDAthreeM}\,dB and PSNR by \nmu{CDAthreeF}\,dB. It also lowers the Dyn.\ PSNR of STG and of the 4DSloMo reconstruction stage by 1.02 and 2.42\,dB. On N3DV, 48 stratified random schedules with our group sizes are 0.39\,dB below ours on average, recovering 69\% of the modular schedule's loss. Our two criteria rank them only weakly, and three of them outperform ours (Sec.~S3).\looseness=-1 %

\paragraph{4DSloMo.}
The full 4DSloMo pipeline, with its own schedule, is 3.04\,dB below OffBeat in Dyn.\ PSNR and 4.35\,dB in PSNR over all six scenes (Table~\ref{tab:main}, Fig.~\ref{fig:slomofull}). Its diffusion stage (about a day per scene) raises the Dyn.\ PSNR of its own reconstruction stage by 2.03\,dB; our run of that stage matches it within 0.09\,dB on average.
 Sec.~S5 relates 4DSloMo's published 33.48\,dB to our protocol.\looseness=-1 %

\subsection{PanopticSports}
\label{sec:exp_pan}
\begin{table}[t]
\centering
\caption{\textbf{Comparison on PanopticSports} ($K{=}12$, test camera 0, test frames 0--59, mean over six sequences). Best in bold, second best underlined. $^\dagger$Exception: per sequence and metric, the better of the modular and OffBeat schedules.\looseness=-1}
\label{tab:pan}
\setlength{\tabcolsep}{4pt}\renewcommand{\arraystretch}{0.95}
\footnotesize
\begin{tabular}{@{}l ccc@{}}
\toprule
Method & Dyn.\ PSNR$\uparrow$ & PSNR$\uparrow$ & LPIPS$\downarrow$\\
\midrule
\multicolumn{4}{@{}l}{\textit{Synchronized capture}} \\
STG~\cite{li2024spacetime} & 22.31 & 26.87 & .140 \\
STG~\cite{li2024spacetime} + interpolation~\cite{li2023amt} & 23.09 & 27.59 & .159 \\
\midrule
\multicolumn{4}{@{}l}{\textit{Staggered capture, OffBeat schedule}} \\
4DSloMo~\cite{chen20254dslomo}, recon.\ stage$^\dagger$ & 23.30 & 24.30 & .310 \\
STG~\cite{li2024spacetime} & \underline{23.86} & \underline{27.76} & \underline{.138} \\
STG~\cite{li2024spacetime}, 2$\times$ iterations & 23.75 & 27.52 & \textbf{.130} \\
\textbf{OffBeat (Ours)} & \textbf{23.98} & \textbf{27.98} & \underline{.138} \\
\bottomrule
\end{tabular}
\end{table}
On PanopticSports, the 27 training cameras form groups of two or three, each triangulating (Sec.~\ref{sec:theory}). Between the first and last synchronized frame (frames 0--48, Sec.~S6), OffBeat outperforms synchronized STG by \nmu{PDSMint}\,dB in Dyn.\ PSNR and is on par in Dyn.\ PSNR and PSNR with STG + interpolation. In this range, synchronized STG is slightly better in LPIPS, and STG + interpolation is better in full-frame and dynamic-region SSIM on every sequence. In Table~\ref{tab:pan}, 4DSloMo's reconstruction stage is \nmu{PDSlomoM}\,dB lower in Dyn.\ PSNR and \nmu{PDSlomoF}\,dB in PSNR (Fig.~\ref{fig:pan_qual}), but better in Dyn.\ PSNR on Softball and Tennis, where OffBeat blurs the arms, bat and racket. At camera~0, the gain comes from staggering: unlike on N3DV, OffBeat is on par with STG on our schedule and its floor variant (Sec.~S6). At the other three test cameras, OffBeat outperforms STG on our schedule on 13 of the 18 sequence--camera pairs in Dyn.\ PSNR. In frames 0--48, against the best synchronized method, it is on par or better in Dyn.\ PSNR and PSNR on 15 and 13 pairs but worse in LPIPS on 7 (Sec.~S6).\looseness=-1 %

\subsection{Analysis}
\label{sec:exp_rule2}
\begin{table}[t]
\centering
\caption{\textbf{Ablation on N3DV} (test frames 100--159, mean over six scenes). $^\ast$Per-image timestamps and no opacity reset (the floor variant). $^\dagger$With STG's unscaled velocity; its static-pixel residual power at $1/T$ is 1.58$\times$ OffBeat's (Fig.~\ref{fig:flicker}).\looseness=-1} %
\label{tab:abl}
\setlength{\tabcolsep}{4pt}\renewcommand{\arraystretch}{0.95}
\footnotesize
\begin{tabular}{@{}l ccc@{}}
\toprule
Variant & Dyn.\ PSNR$\uparrow$ & PSNR$\uparrow$ & LPIPS$\downarrow$\\
\midrule
\multicolumn{4}{@{}l}{\textit{Adding components}} \\
STG~\cite{li2024spacetime}, synchronized & 24.33 & 30.57 & .117 \\
STG~\cite{li2024spacetime}, modular schedule & 26.07 & 30.47 & .125 \\
STG~\cite{li2024spacetime}, OffBeat schedule & 27.09 & 30.85 & .116 \\
\;\;+ lifetime floor & 27.38 & 31.13 & .115 \\
\;\;+ training settings$^\ast$ & 27.73 & 31.12 & .108 \\
\;\;+ window, densification (Ours) & 27.90 & 31.25 & .104 \\
\midrule
\multicolumn{4}{@{}l}{\textit{Replacing one component}} \\
window $\to$ Gaussian$^\dagger$ & 27.91 & 31.30 & .102 \\
densification $\to$ STG's & 27.63 & 31.03 & .111 \\
schedule $\to$ modular & 26.61 & 30.79 & .111 \\
schedule $\to$ synchronized & 23.80 & 30.06 & .118 \\
\bottomrule
\end{tabular}
\end{table}
\paragraph{Ablation study.} %
In Table~\ref{tab:abl}, staggering raises the Dyn.\ PSNR of STG from \nm{NVSM} to 26.07\,dB on the modular schedule and to \nm{NVNM}\,dB on ours. Most of the further gain comes from the lifetime floor (Prop.~\ref{prop:window}, \nmu{NXNfDNM}\,dB) and from per-image timestamps with no opacity reset (\nmu{NXNpDNfM}\,dB). The period window with our densification adds \nmu{NDNpM}\,dB, which comes from Flame Steak and Cook Spinach, whose frames 40--99 we used for validation. %
On the other four scenes, OffBeat is \nma{NODNpM}\,dB below the floor variant in Dyn.\ PSNR but still outperforms the best synchronized method by \nmu{NODSyncM}\,dB and STG on our schedule by \nmu{NODNM}\,dB (Sec.~S6). Under the same floor, Gaussian windows are on par in Dyn.\ PSNR and better in PSNR (\nma{CDAoneF}\,dB) and LPIPS, whereas STG's guided sampling in place of our densification lowers Dyn.\ PSNR by \nmu{CDAtwoM}\,dB. %
In Dyn.\ PSNR, no component helps synchronized capture: the floor variant and our reconstruction differ from STG by $\nm{NXSpDSM}$ and $\nm{NXRsDSM}$\,dB (Sec.~S6).\looseness=-1 %
\paragraph{Flicker at the camera frame rate.} %
On static pixels (Fig.~\ref{fig:flicker}, bottom), STG on our schedule has 3.0--3.7$\times$ the residual power of synchronized STG at the first five harmonics. The lifetime floor alone lowers this to 1.01--1.05$\times$ at harmonics 3--5 and to 2.7$\times$ at the fundamental, which the training settings raise to 5.7$\times$ while improving Dyn.\ PSNR. OffBeat is at 3.4$\times$ at the fundamental, within STG's range, but 1.1$\times$ at the fifth harmonic, and on Flame Salmon lacks STG's single-frame stripes (Fig.~\ref{fig:flicker}, top). Gaussian windows in place of the period window raise the fundamental power on all six scenes, to 1.58$\times$ OffBeat's (Sec.~S7).\looseness=-1 %

\section{Conclusion}
\label{sec:concl}
We presented OffBeat, a 4D capture method that models a staggered rig as a multicoset sampler to follow motion a synchronized rig misses. Under this model, a position-ordered row confounds depth with motion, the calibration alone yields a schedule, and temporal windows weighting all groups equally for every $K$ last at least~$T$, a bound our closed-form period window satisfies. On N3DV, OffBeat outperforms the full 4DSloMo pipeline in dynamic-region and full-frame PSNR on every test frame. Between the first and last synchronized frame, it follows motion better on average than a synchronized rig of equal camera frame rate, which is better at its own frames. On PanopticSports, in the same range, OffBeat is comparable in PSNR to synchronized capture with interpolation. %

\paragraph{Limitations and Future Work.}
We simulate staggered capture by subsampling 30\,fps video, and the schedule needs calibration before capture. On 4DSloMo's real staggered recording, we test only the lifetime floor, in its reconstruction stage, and find no reliable gain. In that stage, the period window as used on STG lowers dynamic-region PSNR on an N3DV validation scene (Sec.~S5). OffBeat is more sensitive than synchronized STG to simulated trigger-time and photometric errors, which a real rig adds (Secs.~S7 and~S8). Estimating trigger times and gains during reconstruction could lower this sensitivity. Content shorter than $T$ cannot be represented, and the fastest motion between a group's firings stays blurred. OffBeat fails on Juggle at one PanopticSports camera (Sec.~S6).\looseness=-1 %

\section*{LLM Usage Statement}
The authors chose the research problem and the core method and directed the use of Claude (Anthropic) through the Claude Code agent. Claude wrote and reviewed the experiment code, ran and monitored the experiments, computed the statistics, kept an append-only experiment log, and drafted the text and figures, including this statement. The simulated views of the velocity test in Sec.~S4 were generated with code written by Claude. The authors approved each experiment plan before it was run, read every result, and edited the text. Gemini (Google) was used only to critique drafts. The authors are fully responsible for the content.

{\small
\bibliographystyle{ieeenat_fullname}
\bibliography{references}

\begin{thebibliography}{53}
\providecommand{\natexlab}[1]{#1}
\providecommand{\url}[1]{\texttt{#1}}
\expandafter\ifx\csname urlstyle\endcsname\relax
  \providecommand{\doi}[1]{doi: #1}\else
  \providecommand{\doi}{doi: \begingroup \urlstyle{rm}\Url}\fi

\bibitem[Agrawal et~al.(2010)Agrawal, Gupta, Veeraraghavan, and
  Narasimhan]{agrawal2010optimal}
Amit Agrawal, Mohit Gupta, Ashok Veeraraghavan, and Srinivasa~G. Narasimhan.
\newblock Optimal coded sampling for temporal super-resolution.
\newblock In \emph{CVPR}, 2010.

\bibitem[Albl et~al.(2017)Albl, Kukelova, Fitzgibbon, Heller, Smid, and
  Pajdla]{albl2017two}
Cenek Albl, Zuzana Kukelova, Andrew Fitzgibbon, Jan Heller, Matej Smid, and
  Tomas Pajdla.
\newblock On the two-view geometry of unsynchronized cameras.
\newblock In \emph{CVPR}, 2017.

\bibitem[Avidan and Shashua(2000)]{avidan2000trajectory}
Shai Avidan and Amnon Shashua.
\newblock Trajectory triangulation: 3{D} reconstruction of moving points from a
  monocular image sequence.
\newblock \emph{IEEE Transactions on Pattern Analysis and Machine
  Intelligence}, 22\penalty0 (4):\penalty0 348--357, 2000.

\bibitem[Blu et~al.(2001)Blu, Th{\'e}venaz, and Unser]{blu2001moms}
Thierry Blu, Philippe Th{\'e}venaz, and Michael Unser.
\newblock {MOMS}: Maximal-order interpolation of minimal support.
\newblock \emph{IEEE Transactions on Image Processing}, 10\penalty0
  (7):\penalty0 1069--1080, 2001.

\bibitem[Boyd et~al.(2017)Boyd, Schiebinger, and Recht]{boyd2017adcg}
Nicholas Boyd, Geoffrey Schiebinger, and Benjamin Recht.
\newblock The alternating descent conditional gradient method for sparse
  inverse problems.
\newblock \emph{SIAM Journal on Optimization}, 27\penalty0 (2):\penalty0
  616--639, 2017.

\bibitem[Bredies and Pikkarainen(2013)]{bredies2013inverse}
Kristian Bredies and Hanna~Katriina Pikkarainen.
\newblock Inverse problems in spaces of measures.
\newblock \emph{ESAIM: Control, Optimisation and Calculus of Variations},
  19\penalty0 (1):\penalty0 190--218, 2013.

\bibitem[Breuer et~al.(2005)Breuer, Blaimer, Heidemann, Mueller, Griswold, and
  Jakob]{breuer2005caipirinha}
Felix~A. Breuer, Martin Blaimer, Robin~M. Heidemann, Matthias~F. Mueller,
  Mark~A. Griswold, and Peter~M. Jakob.
\newblock Controlled aliasing in parallel imaging results in higher
  acceleration ({CAIPIRINHA}) for multi-slice imaging.
\newblock \emph{Magnetic Resonance in Medicine}, 53\penalty0 (3):\penalty0
  684--691, 2005.

\bibitem[Caspi and Irani(2002)]{caspi2002spatio}
Yaron Caspi and Michal Irani.
\newblock Spatio-temporal alignment of sequences.
\newblock \emph{IEEE Transactions on Pattern Analysis and Machine
  Intelligence}, 24\penalty0 (11):\penalty0 1409--1424, 2002.

\bibitem[Chen et~al.(2025)Chen, Guo, Yang, Ding, Yu, Gu, and
  Xue]{chen20254dslomo}
Yutian Chen, Shi Guo, Tianshuo Yang, Lihe Ding, Xiuyuan Yu, Jinwei Gu, and
  Tianfan Xue.
\newblock {4DSloMo}: {4D} reconstruction for high speed scene with asynchronous
  capture.
\newblock In \emph{SIGGRAPH Asia}, 2025.
\newblock arXiv:2507.05163.

\bibitem[Gao et~al.(2022)Gao, Li, Tulsiani, Russell, and
  Kanazawa]{gao2022monocular}
Hang Gao, Ruilong Li, Shubham Tulsiani, Bryan Russell, and Angjoo Kanazawa.
\newblock Monocular dynamic view synthesis: A reality check.
\newblock In \emph{NeurIPS}, 2022.

\bibitem[Huang et~al.(2025)Huang, Guan, Shang, and Yu]{huang2025trajectory}
Huayu Huang, Banglei Guan, Yang Shang, and Qifeng Yu.
\newblock {3D} trajectory reconstruction of moving points based on asynchronous
  cameras.
\newblock \emph{arXiv preprint arXiv:2506.00541}, 2025.

\bibitem[Joo et~al.(2015)Joo, Liu, Tan, Gui, Nabbe, Matthews, Kanade, Nobuhara,
  and Sheikh]{joo2015panoptic}
Hanbyul Joo, Hao Liu, Lei Tan, Lin Gui, Bart Nabbe, Iain Matthews, Takeo
  Kanade, Shohei Nobuhara, and Yaser Sheikh.
\newblock {Panoptic Studio}: A massively multiview system for social motion
  capture.
\newblock In \emph{ICCV}, 2015.

\bibitem[Kerbl et~al.(2023)Kerbl, Kopanas, Leimk{\"u}hler, and
  Drettakis]{kerbl20233d}
Bernhard Kerbl, Georgios Kopanas, Thomas Leimk{\"u}hler, and George Drettakis.
\newblock 3{D} {Gaussian} splatting for real-time radiance field rendering.
\newblock \emph{ACM Transactions on Graphics}, 42\penalty0 (4), 2023.

\bibitem[Kiefer(1974)]{kiefer1974general}
Jack Kiefer.
\newblock General equivalence theory for optimum designs (approximate theory).
\newblock \emph{The Annals of Statistics}, 2\penalty0 (5):\penalty0 849--879,
  1974.

\bibitem[Kim et~al.(2024)Kim, Bae, Yun, Lee, Bang, and Uh]{kim2024sync}
Seoha Kim, Jeongmin Bae, Youngsik Yun, Hahyun Lee, Gun Bang, and Youngjung Uh.
\newblock Sync-{NeRF}: Generalizing dynamic {NeRF}s to unsynchronized videos.
\newblock In \emph{AAAI}, 2024.

\bibitem[Kim et~al.(2026)Kim, Kim, Yoo, and Yun]{kim2026spin}
Seung-gyeom Kim, Areum Kim, Yongjae Yoo, and Sukmin Yun.
\newblock Implicit {4D} {G}aussian splatting for fast motion with large
  inter-frame displacements.
\newblock In \emph{ICLR}, 2026.

\bibitem[K{\"u}nsch(1989)]{kunsch1989jackknife}
Hans~R. K{\"u}nsch.
\newblock The jackknife and the bootstrap for general stationary observations.
\newblock \emph{The Annals of Statistics}, 17\penalty0 (3):\penalty0
  1217--1241, 1989.

\bibitem[Lee et~al.(2024)Lee, Won, Jung, Bae, and Jeon]{lee2024fully}
Junoh Lee, Changyeon Won, Hyunjun Jung, Inhwan Bae, and Hae-Gon Jeon.
\newblock Fully explicit dynamic {G}aussian splatting.
\newblock In \emph{NeurIPS}, 2024.

\bibitem[Lee et~al.(2025)Lee, Huang, Gehlot, Choi, Su, and
  Manocha]{synctrack4d2025}
Yonghan Lee, Tsung-Wei Huang, Shiv Gehlot, Jaehoon Choi, Guan-Ming Su, and
  Dinesh Manocha.
\newblock {SyncTrack4D}: Cross-video motion alignment and video synchronization
  for multi-video {4D} {Gaussian} splatting, 2025.
\newblock arXiv:2512.04315.

\bibitem[Li et~al.(2022)Li, Slavcheva, Zollh{\"o}fer, Green, Lassner, Kim,
  Schmidt, Lovegrove, Goesele, Newcombe, and Lv]{li2022neural}
Tianye Li, Mira Slavcheva, Michael Zollh{\"o}fer, Simon Green, Christoph
  Lassner, Changil Kim, Tanner Schmidt, Steven Lovegrove, Michael Goesele,
  Richard Newcombe, and Zhaoyang Lv.
\newblock Neural 3{D} video synthesis from multi-view video.
\newblock In \emph{CVPR}, 2022.

\bibitem[Li et~al.(2023)Li, Zhu, Han, Hou, Guo, and Cheng]{li2023amt}
Zhen Li, Zuo-Liang Zhu, Ling-Hao Han, Qibin Hou, Chun-Le Guo, and Ming-Ming
  Cheng.
\newblock {AMT}: All-pairs multi-field transforms for efficient frame
  interpolation.
\newblock In \emph{CVPR}, 2023.

\bibitem[Li et~al.(2024)Li, Chen, Li, and Xu]{li2024spacetime}
Zhan Li, Zhang Chen, Zhong Li, and Yi Xu.
\newblock Spacetime {Gaussian} feature splatting for real-time dynamic view
  synthesis.
\newblock In \emph{CVPR}, 2024.

\bibitem[Liao et~al.(2026)Liao, Zhang, Tu, Wang, Gao, Zhang, and
  Liu]{liao2026sharptimegs}
Zhanfeng Liao, Jiajun Zhang, Hanzhang Tu, Zhixi Wang, Yunqi Gao, Hongwen Zhang,
  and Yebin Liu.
\newblock {SharpTimeGS}: Sharp and stable dynamic {Gaussian} splatting via
  lifespan modulation.
\newblock In \emph{CVPR}, 2026.

\bibitem[Luiten et~al.(2024)Luiten, Kopanas, Leibe, and
  Ramanan]{luiten2024dynamic}
Jonathon Luiten, Georgios Kopanas, Bastian Leibe, and Deva Ramanan.
\newblock Dynamic {3D} {G}aussians: Tracking by persistent dynamic view
  synthesis.
\newblock In \emph{3DV}, 2024.

\bibitem[Lyu et~al.(2025)Lyu, Cheng, Kang, and Chen]{lyu2024resgs}
Yanzhe Lyu, Kai Cheng, Xin Kang, and Xuejin Chen.
\newblock {ResGS}: Residual densification of {3D} {Gaussian} for efficient
  detail recovery.
\newblock In \emph{ICCV}, 2025.

\bibitem[Nardone and Aidala(1981)]{nardone1981observability}
Steven~C. Nardone and Vincent~J. Aidala.
\newblock Observability criteria for bearings-only target motion analysis.
\newblock \emph{IEEE Transactions on Aerospace and Electronic Systems},
  AES-17\penalty0 (2):\penalty0 162--166, 1981.

\bibitem[Novikov et~al.(2026)Novikov, Vaknin, Tumanyan, and
  Sheinin]{novikov2026color}
David Novikov, Eilon Vaknin, Narek Tumanyan, and Mark Sheinin.
\newblock Color-encoded illumination for high-speed volumetric scene
  reconstruction.
\newblock In \emph{CVPR}, 2026.

\bibitem[Papoulis(1977)]{papoulis1977generalized}
Athanasios Papoulis.
\newblock Generalized sampling expansion.
\newblock \emph{IEEE Transactions on Circuits and Systems}, 24\penalty0
  (11):\penalty0 652--654, 1977.

\bibitem[Park et~al.(2010)Park, Shiratori, Matthews, and Sheikh]{park20103d}
Hyun~Soo Park, Takaaki Shiratori, Iain Matthews, and Yaser Sheikh.
\newblock 3{D} reconstruction of a moving point from a series of 2{D}
  projections.
\newblock In \emph{ECCV}, 2010.

\bibitem[Pruessmann et~al.(1999)Pruessmann, Weiger, Scheidegger, and
  Boesiger]{pruessmann1999sense}
Klaas~P. Pruessmann, Markus Weiger, Markus~B. Scheidegger, and Peter Boesiger.
\newblock {SENSE}: Sensitivity encoding for fast {MRI}.
\newblock \emph{Magnetic Resonance in Medicine}, 42\penalty0 (5):\penalty0
  952--962, 1999.

\bibitem[Pukelsheim(2006)]{pukelsheim2006optimal}
Friedrich Pukelsheim.
\newblock \emph{Optimal Design of Experiments}.
\newblock SIAM, 2006.
\newblock Reprint of the 1993 Wiley edition.

\bibitem[Reda et~al.(2022)Reda, Kontkanen, Tabellion, Sun, Pantofaru, and
  Curless]{reda2022film}
Fitsum Reda, Janne Kontkanen, Eric Tabellion, Deqing Sun, Caroline Pantofaru,
  and Brian Curless.
\newblock {FILM}: Frame interpolation for large motion.
\newblock In \emph{ECCV}, 2022.

\bibitem[Reed et~al.(1983)Reed, Gagliardi, and Shao]{reed1983tbd}
Irving~S. Reed, Robert~M. Gagliardi, and Huey~M. Shao.
\newblock Application of three-dimensional filtering to moving target
  detection.
\newblock \emph{IEEE Transactions on Aerospace and Electronic Systems},
  19\penalty0 (6):\penalty0 898--905, 1983.

\bibitem[Roh et~al.(2026)Roh, Lee, Jeong, Lee, Lee, Park, and
  Park]{roh2026dusting}
Minhyeok Roh, Minseo Lee, Jun~Young Jeong, Byeonghyeon Lee, Lucas~Yunkyu Lee,
  Jaesik Park, and Eunbyung Park.
\newblock Dusting the phantom: Gaussian allocation for fading dynamic content.
\newblock \emph{arXiv preprint arXiv:2610.21302}, 2026.

\bibitem[Rota~Bul{\`o} et~al.(2024)Rota~Bul{\`o}, Porzi, and
  Kontschieder]{rotabulo2024revising}
Samuel Rota~Bul{\`o}, Lorenzo Porzi, and Peter Kontschieder.
\newblock Revising densification in {Gaussian} splatting.
\newblock In \emph{ECCV}, 2024.

\bibitem[Sandu et~al.(2026)Sandu, Pathak, and
  Soundararajan]{sandu2026temporally}
Vikram Sandu, Mayurdeep Pathak, and Rajiv Soundararajan.
\newblock Temporally aware densification for dynamic {3D} {Gaussian} splatting.
\newblock In \emph{ECCV}, 2026.

\bibitem[Shechtman et~al.(2005)Shechtman, Caspi, and Irani]{shechtman2005space}
Eli Shechtman, Yaron Caspi, and Michal Irani.
\newblock Space-time super-resolution.
\newblock \emph{IEEE Transactions on Pattern Analysis and Machine
  Intelligence}, 27\penalty0 (4):\penalty0 531--545, 2005.

\bibitem[Strang and Fix(1973)]{strang1973fourier}
Gilbert Strang and George Fix.
\newblock A {Fourier} analysis of the finite element variational method.
\newblock In \emph{Constructive Aspects of Functional Analysis}, pages
  793--840. Cremonese, 1973.

\bibitem[Su et~al.(2025)Su, Feng, Gehrig, Jiang, Gao, Lagorce, and
  Kneip]{su2025linear}
Hang Su, Yunlong Feng, Daniel Gehrig, Panfeng Jiang, Ling Gao, Xavier Lagorce,
  and Laurent Kneip.
\newblock A linear {N}-point solver for structure and motion from asynchronous
  tracks.
\newblock In \emph{ICCV}, 2025.

\bibitem[Tsao et~al.(2003)Tsao, Boesiger, and Pruessmann]{tsao2003kt}
Jeffrey Tsao, Peter Boesiger, and Klaas~P. Pruessmann.
\newblock k-t {BLAST} and k-t {SENSE}: Dynamic {MRI} with high frame rate
  exploiting spatiotemporal correlations.
\newblock \emph{Magnetic Resonance in Medicine}, 50\penalty0 (5):\penalty0
  1031--1042, 2003.

\bibitem[Venkataramani and Bresler(2001)]{venkataramani2001optimal}
Raman Venkataramani and Yoram Bresler.
\newblock Optimal sub-{N}yquist nonuniform sampling and reconstruction for
  multiband signals.
\newblock \emph{IEEE Transactions on Signal Processing}, 49\penalty0
  (10):\penalty0 2301--2313, 2001.

\bibitem[Vo et~al.(2016)Vo, Narasimhan, and Sheikh]{vo2016spatiotemporal}
Minh Vo, Srinivasa~G. Narasimhan, and Yaser Sheikh.
\newblock Spatiotemporal bundle adjustment for dynamic 3{D} reconstruction.
\newblock In \emph{CVPR}, 2016.

\bibitem[Wang et~al.(2025{\natexlab{a}})Wang, Wang, Wang, Mohan, Fan, Wu, Cai,
  Yeh, Wang, Liu, and Ranjan]{wang2025steepgs}
Peihao Wang, Yuehao Wang, Dilin Wang, Sreyas Mohan, Zhiwen Fan, Lemeng Wu,
  Ruisi Cai, Yu-Ying Yeh, Zhangyang Wang, Qiang Liu, and Rakesh Ranjan.
\newblock Steepest descent density control for compact {3D} {Gaussian}
  splatting.
\newblock In \emph{CVPR}, 2025{\natexlab{a}}.

\bibitem[Wang et~al.(2026)Wang, Ma, Shen, Wang, and Sander]{wang2026retimegs}
Xuezhen Wang, Li Ma, Yulin Shen, Zeyu Wang, and Pedro~V. Sander.
\newblock {RetimeGS}: Continuous-time reconstruction of {4D} {Gaussian}
  splatting.
\newblock In \emph{CVPR}, 2026.
\newblock arXiv:2603.13783.

\bibitem[Wang et~al.(2025{\natexlab{b}})Wang, Yang, Xu, Sun, Zhang, Chen, Bao,
  Peng, and Zhou]{wang2025freetimegs}
Yifan Wang, Peishan Yang, Zhen Xu, Jiaming Sun, Zhanhua Zhang, Yong Chen, Hujun
  Bao, Sida Peng, and Xiaowei Zhou.
\newblock {FreeTimeGS}: Free {Gaussian} primitives at anytime anywhere for
  dynamic scene reconstruction.
\newblock In \emph{CVPR}, 2025{\natexlab{b}}.

\bibitem[Wilburn et~al.(2005)Wilburn, Joshi, Vaish, Talvala, Antunez, Barth,
  Adams, Horowitz, and Levoy]{wilburn2005high}
Bennett Wilburn, Neel Joshi, Vaibhav Vaish, Eino-Ville Talvala, Emilio Antunez,
  Adam Barth, Andrew Adams, Mark Horowitz, and Marc Levoy.
\newblock High performance imaging using large camera arrays.
\newblock \emph{ACM Transactions on Graphics}, 24\penalty0 (3):\penalty0
  765--776, 2005.

\bibitem[Wu et~al.(2024)Wu, Yi, Fang, Xie, Zhang, Wei, Liu, Tian, and
  Wang]{wu20244d}
Guanjun Wu, Taoran Yi, Jiemin Fang, Lingxi Xie, Xiaopeng Zhang, Wei Wei, Wenyu
  Liu, Qi Tian, and Xinggang Wang.
\newblock 4{D} {Gaussian} splatting for real-time dynamic scene rendering.
\newblock In \emph{CVPR}, 2024.

\bibitem[Xu et~al.(2026)Xu, Zhou, Liu, Xue, Pan, Wang, and Wang]{xu2026unsync}
Zhixin Xu, Hengyu Zhou, Yuan Liu, Wenhan Xue, Hao Pan, Wenping Wang, and Bin
  Wang.
\newblock Dynamic {Gaussian} scene reconstruction from unsynchronized videos.
\newblock In \emph{AAAI}, 2026.
\newblock arXiv:2511.11175.

\bibitem[Yang et~al.(2024)Yang, Yang, Pan, and Zhang]{yang2024real}
Zeyu Yang, Hongye Yang, Zijie Pan, and Li Zhang.
\newblock Real-time photorealistic dynamic scene representation and rendering
  with {4D} {G}aussian splatting.
\newblock In \emph{ICLR}, 2024.

\bibitem[Yu et~al.(2024)Yu, Chen, Huang, Sattler, and Geiger]{yu2024mip}
Zehao Yu, Anpei Chen, Binbin Huang, Torsten Sattler, and Andreas Geiger.
\newblock {Mip-Splatting}: Alias-free {3D} {Gaussian} splatting.
\newblock In \emph{CVPR}, 2024.

\bibitem[Zhang et~al.(2018)Zhang, Isola, Efros, Shechtman, and
  Wang]{zhang2018unreasonable}
Richard Zhang, Phillip Isola, Alexei~A. Efros, Eli Shechtman, and Oliver Wang.
\newblock The unreasonable effectiveness of deep features as a perceptual
  metric.
\newblock In \emph{CVPR}, 2018.

\bibitem[Zheng et~al.(2015)Zheng, Ji, Dunn, and Frahm]{zheng2015sparse}
Enliang Zheng, Dinghuang Ji, Enrique Dunn, and Jan-Michael Frahm.
\newblock Sparse dynamic 3{D} reconstruction from unsynchronized videos.
\newblock In \emph{ICCV}, 2015.

\bibitem[Zhu et~al.(2026)Zhu, Zhu, Zhang, Li, Ma, and Cao]{zhu2026astra}
Junyu Zhu, Hao Zhu, Xinzhuo Zhang, Hongdong Li, Zhan Ma, and Xun Cao.
\newblock {ASTRA}: Asynchronous spatio-temporal reconstruction via trajectory
  alignment.
\newblock \emph{arXiv preprint arXiv:2608.02006}, 2026.

\end{thebibliography}
}
\end{document}